\documentclass{article}
\usepackage{spconf,amsmath,graphicx,hyperref}
\usepackage{algorithm}
\usepackage{algpseudocode}
\usepackage{amssymb}
\usepackage{booktabs}
\usepackage{bm}
\usepackage{enumitem}
\usepackage{graphicx}
\usepackage{xcolor}

\usepackage[caption=false,font=normalsize,labelfont=footnotesize,textfont=sf]{subfig}
\def\BibTeX{{\rm B\kern-.05em{\sc i\kern-.025em b}\kern-.08em
    T\kern-.1667em\lower.7ex\hbox{E}\kern-.125emX}}
\usepackage{tikz}
\newcommand\copyrighttext{%
  \footnotesize \textcopyright 2026 IEEE. Personal use of this material is permitted. Permission from IEEE must be obtained for all other uses, in any current or future media, including reprinting/republishing this material for advertising or promotional purposes, creating new collective works, for resale or redistribution to servers or lists, or reuse of any copyrighted component of this work in other works.}
\newcommand\copyrightstatement{%
\begin{tikzpicture}[remember picture,overlay]
\node[anchor=south,yshift=10pt] at (current page.south) {\fbox{\parbox{\dimexpr\textwidth-\fboxsep-\fboxrule\relax}{\copyrighttext}}};
\end{tikzpicture}%
}

\title{Differentially Private Clustered Federated Learning with Privacy-Preserving Initialization and Normality-Driven Aggregation}
\name{Jie Xu\textsuperscript{*}, Haaris Mehmood\textsuperscript{*}, Rogier Van Dalen\textsuperscript{\textdagger}, Karthikeyan Saravanan\textsuperscript{*}, and Mete Ozay\textsuperscript{*}
\address{\textsuperscript{*}Samsung R\&D Institute UK (SRUK), \textsuperscript{\textdagger}Samsung AI Centre Cambridge\\}
}
\begin{document}
\ninept
\maketitle
%

\copyrightstatement

\begin{abstract}
Federated learning (FL) enables training of a global model while keeping raw data on end‑devices. Despite this, FL has shown to leak private user information and thus in practice, it is often coupled with methods such as differential privacy (DP) and secure vector sum to provide formal privacy guarantees to its participants. In realistic cross‑device deployments, the data are highly heterogeneous, so vanilla federated learning converges slowly and generalizes poorly. Clustered federated learning (CFL) mitigates this by segregating users into clusters, leading to lower intra-cluster data heterogeneity. Nevertheless, coupling CFL with DP remains challenging: the injected DP noise makes individual client updates excessively noisy, and the server is unable to initialize cluster centroids with the less noisy aggregated updates.
To address this challenge, we propose PINA, a two‑stage framework that first lets each client fine‑tune a lightweight low-rank adaptation (LoRA) adapter and privately share a compressed sketch of the update.
The server leverages these sketches to construct robust cluster centroids.
In the second stage, PINA introduces a normality-driven aggregation mechanism that improves convergence and robustness.
Our method retains the benefits of clustered FL while providing formal privacy guarantees against an untrusted server. Extensive evaluations show that our proposed method outperforms state‑of‑the‑art DP‑FL algorithms by an average of 2.9\% in accuracy for privacy budgets (\(\epsilon\in\{2,8\}\)).
\end{abstract}


\section{Introduction}
\label{sec:intro}

Federated learning (FL) enables a distributed group of edge devices to collaboratively train a shared model while keeping raw user data on-device \cite{fedavg}. Despite this, the exchanged gradients or model updates can reveal statistical fingerprints that compromise user privacy \cite{dp-fedavg}.  Differential privacy (DP) \cite{dp} protects against such inferences by injecting calibrated randomness into the updates, yielding a rigorous privacy guarantee. Local DP (LDP) \cite{ldp} requires client updates to be privatized at the edge and therefore protects against a malicious server. However, the higher privacy benefits of LDP leads to noisier updates and drop in model performance. The alternative solution is to add noise to the sum of contributions which is known as central DP (CDP) \cite{dp}, and leverage secure sum protocols \cite{secagg, ddp_secure_sum} to ensure that the server only has access to the aggregated updates and is prevented from viewing individual contributions.

A second fundamental challenge in FL is data heterogeneity: client data are often non‑independent and identically distributed (non‑IID) \cite{fl_heterogeneity}. Under such conditions, standard FL algorithms such as FedAvg \cite{fedavg} can converge slowly and yield suboptimal performance \cite{fl_heterogeneity}. The situation deteriorates further when DP is imposed: the added DP noise amplifies the impact of data skew, exacerbating the performance degradation caused by non-IID data  \cite{dp-fedprox,dp-scaffold,dpfl_heterogeneity}.

Clustered federated learning (CFL) has been recently proposed to address data heterogeneity in FL \cite{ifca,cfl}. Instead of fitting a single global model, CFL simultaneously learns a set of cluster‑specific models and dynamically assigns clients to the cluster whose data distribution best matches their local statistics. Clients that share a similar distribution form a cluster, making the data within each cluster less non-IID and enabling each cluster to converge faster and to a higher accuracy. Existing CFL algorithms, such as IFCA \cite{ifca}, rely on privileged server data that resembles users’ data to initialize cluster models. An alternative approach employs random restarts to initialize cluster models arbitrarily and repeat training until a stable clustering structure emerges. However, both strategies are problematic in a DP setting. The former requires access to non‑public data at the server, which compromises privacy and undermines practicality. The latter consumes additional privacy budget, resulting in noisier client updates. Moreover, a naïve application of LDP in CFL introduces excessive noise, which distorts the intrinsic cluster structure and severely degrades model performance. On the other hand, simply applying CDP to CFL requires the server to be fully trusted, which is impractical in adversarial settings. To protect against an untrusted server, secure sum protocols such as SecAgg \cite{secagg} are often deployed to ensure the server only has access to the aggregated contributions. However, such mechanisms prevent the server from accessing individual client updates, which are necessary for initializing cluster models in the first place.

In this paper, we introduce Clustered Federated Learning with \textbf{P}rivacy-preserving \textbf{I}nitialization and \textbf{N}ormality-driven \textbf{A}ggregation (PINA), a novel CFL framework that ensures DP without requiring privileged server data or random restarts. PINA uses privacy-preserving sketches of client updates to construct robust initial cluster prototypes, enabling dynamic client assignment to the nearest cluster. In subsequent training rounds, PINA employs a normality-driven aggregation mechanism that adaptively mitigates the impact of imbalanced client contributions. We believe that these advancements brings clustered FL closer to practical adoption, providing realistic privacy guarantees. The contributions of this work are summarized as follows: 

\begin{itemize}[leftmargin=*]
\itemsep0em
    \item We propose PINA, a clustered FL algorithm that jointly addresses data heterogeneity and privacy, requiring neither privileged server data nor random restarts.
    \item We devise a privacy‑preserving initialization scheme that produces accurate cluster prototypes directly from privatized client sketches.
    \item We introduce a normality‑driven aggregation step that restores the magnitude of the aggregated updates, improving robustness against imbalanced client participation.
    \item Our extensive experiments demonstrate that PINA consistently outperforms existing DP‑FL methods on non-IID data by an average of $2.9\%$ in test accuracy, particularly in more realistic, naturally non-IID environments.
\end{itemize}

\section{Preliminary}
\label{sec:preliminary}

\subsection{Federated Learning (FL)}

\textbf{Overview of FL}: At the start of each communication round $t$, a global model $W^t$ is provided by the server and a randomly sampled user set $\mathcal{K}^t$ is constructed. Each user $k \in \mathcal{K}^t$ trains the model locally to obtain $W^t_k$ and shares the model difference ${\Delta^t_k = W^t_k - W^t}$ back to the server. The server aggregates the updates ${W^{t+1} = W^t + \frac{1}{|\mathcal{K}^t|} \sum_{k \in \mathcal{K}^t} \Delta^t_k}$, and then proceeds to the next round.

\subsection{Differential Privacy}
Differential privacy (DP) provides a formal definition to quantify the amount of private information an algorithm leaks regarding its input data. Formally, DP is defined as follows:

\noindent \textbf{Definition (Differential Privacy \cite{dp})} \textit{A randomized mechanism $\mathcal{M}: \mathcal{D} \rightarrow \mathcal{R}$ satisfies ($\epsilon$,$\delta$)-DP if for any pair of adjacent datasets $\mathcal{D}$ and $\mathcal{D}^{\prime}$, and any subset of outputs $\mathcal{S} \subseteq \mathcal{R}$, we have}
\begin{equation}
\label{eqn_dp}
\text{Pr}[\mathcal{M}(\mathcal{D}) \in \mathcal{S}] \leq \exp(\epsilon) \cdot \text{Pr}[\mathcal{M}(\mathcal{D^{\prime}}) \in \mathcal{S}] + \delta,
\end{equation}
where ($\epsilon$,$\delta$) is known as the privacy budget.

Following \cite{dp-fedavg}, we use Gaussian mechanism \cite{gaussian_mechanism} with noise $\mathcal{N}(0, I\sigma^2)$ to achieve DP where $\sigma = z S$. Here, $S$ denotes a predefined threshold for clipping client updates and $z$ denotes the noise multiplier. We compute $z$ using moments accountant \cite{moments_accountant} with R\'enyi Differential Privacy (RDP) \cite{rdp} for a tight composition bound. For any $\alpha \in (1,\infty)$ and $\epsilon > 0$, a randomized mechanism $\mathcal{M}$ satisfies $(\alpha, \epsilon^{\prime})$-RDP if for all neighboring datasets $D$ and $D^{\prime}$, we have:

\noindent \textbf{Definition (R\'enyi Differential Privacy \cite{rdp})} \textit{For any $\alpha \in (1,\infty)$ and $\epsilon > 0$, a randomized mechanism $\mathcal{M}$ satisfies $(\alpha, \epsilon^{\prime})$-RDP if for all neighboring datasets $D$ and $D^{\prime}$, we have}
\begin{equation}
\label{eqn_rdp}
D_{\alpha}(\mathcal{M}(D) \vert \vert \mathcal{M}(D^{\prime})) \triangleq \frac{1}{\alpha - 1} \log \mathbb{E} \left( \frac{D(x)}{D^{\prime}(x)} \right)^{\alpha} \leq \epsilon^{\prime}.
\end{equation}

To convert $(\alpha, \epsilon^{\prime})$-RDP back to the standard $(\epsilon, \delta)$-DP framework, we adopt the established conversion method outlined in \cite{rdp_accountant}. Specifically, a randomised mechanism $\mathcal{M}$ that satisfies $(\alpha, \epsilon^{\prime})$-RDP also satisfies $(\epsilon, \delta)$-DP with
\begin{equation}
\label{eqn_rdp_dp}
\epsilon = \epsilon^{\prime} + \log(\frac{\alpha - 1}{\alpha}) - \frac{\log \alpha + \log \delta}{\alpha - 1},
\end{equation}
for any $0 < \delta < 1$.

For CDP with secure sum, we additionally combine Gaussian mechanism with privacy amplification via sampling \cite{privacy_amplification_via_subsampling_1} by $\sigma = \frac{z S}{|\mathcal{K}^t|}$, achieving a significantly reduced noise level compared to LDP.

\subsection{LoRA}
Low-Rank Adaptation (LoRA) \cite{lora} is a parameter-efficient fine-tuning (PEFT) method for transformer-based pre-trained models. Instead of training the entire weight matrix, it freezes the pre-trained weights $W_0 \in \mathbb R^{d \times d^\prime}$ and introduces new trainable low-rank decomposition matrices $B$ and $A$ as follows:
\begin{equation}
\label{eqn_lora}
W_0 + \Delta W = W_0 + B A,
\end{equation}
where $B \in \mathbb R^{d \times r}$ is initialized to zeros, $A \in \mathbb R^{r \times d^\prime}$ follows random Gaussian initialization and $r \ll \min(d, d^\prime)$. This effectively reduces the number of trainable parameters by an order of $O(r/\min(d, d^\prime))$.

\subsection{Clustered Federated Learning}
Clustered FL algorithms group clients with similar distributions together. Thereby, clients within the same cluster suffer less from data heterogeneity and train via FL more effectively.
Several works propose clustered FL methods, including CFL \cite{cfl}, IFCA\cite{ifca} and PACFL\cite{pacfl}, which use techniques such as cosine similarity, empirical loss and singular value decomposition to assign clients to clusters and train models. Clients typically respond with their cluster ID and trained model.

To our knowledge, \cite{ifca_ldp} and \cite{r-dpcfl} are the only existing works that add DP to clustered FL. However, \cite{ifca_ldp} uses LDP to privatize user updates throughout the entire training process which significantly degrades the model's utility, making it infeasible for training large models \cite{fl_large_model}. Meanwhile, \cite{r-dpcfl} provides sample-level privacy instead of user-level privacy which is a weaker form of privacy protection than the latter. It's also been pointed out that the loss-based clustering in \cite{r-dpcfl} breaks the sample-level privacy guarantees by leaking more information than allowed. In contrast, works including \cite{sacfl, clusterguard} add secure aggregation to clustered FL without incorporating DP constraints, relying on a fully trusted server.

\section{Our Method: PINA}
\label{sec:method}

\subsection{Overview}

\begin{figure}[!t]
\centering
\includegraphics[width=1.0\linewidth]{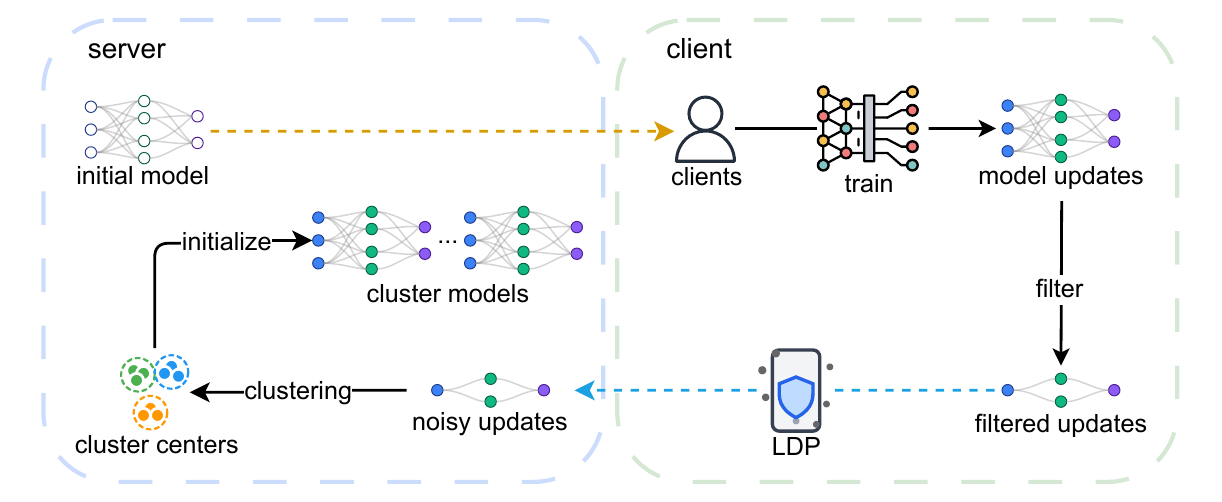}
\vspace*{-20pt}
\caption{Stage 1 of \textit{PINA}. Global cluster models are initialized in a privacy-preserving manner.}
\label{fig_initialization}
\vspace*{-5pt}
\end{figure}

Our  proposed  method  PINA  consists  of  two  stages:  (1) Cluster Model Initialization and (2) Clustered Model Training. In (1), we privately initialize cluster models from user updates; and in (2) we perform cluster identification and model training in a federated setting, privately updating global cluster models. The workflow of PINA is outlined in Algorithm~\ref{alg_pina}.

\begin{figure}[!t]
\vspace*{-5pt}
\begin{algorithm}[H]
\caption{PINA}
\small
\begin{algorithmic}[1]
\State {\textsc{Server}}$ $
\State \hspace{0.5cm}\textbf{parameters: } \#rounds (initialization): $ T^{in} $, 
\#rounds (training): $ T^{tr} $,
a set of users: $ \mathcal{K} $,
sampling rate: $ q \in (0,1] $,
noise multiplier: $ z $,
clipping threshold: $ S^{in}, S $,
number of clusters: $ C $
\State \hspace{0.5cm}\textbf{for} each round $ t = 1, 2, \ldots, T^{in} + T^{tr} $ \textbf{do}
\State \hspace{1.0cm}Sample a subset $\mathcal{K}^t \subseteq \mathcal{K}$ of users uniformly at random
\State \hspace{1.0cm}with probability $q$.
\State \hspace{1.0cm}\textbf{if} $ t \leq T^{in} $
\State \hspace{1.5cm}$ \text{Server} \leftarrow$ {\textsc{Train}}$(t, C, W_0, z, S^{in})$ \Comment{Equations \ref{eqn_init_clip_norm}}
\State \hspace{1.0cm}\textbf{else}
\State \hspace{1.5cm}$W^{t+1} \leftarrow W^t + \frac1{\lvert \mathcal{K}^t \rvert} \textsc{SecureSumDP} \big($
\State \hspace{1.5cm}$\{ \textsc{Train}(t, C, W, z, S)\}_{k \in \mathcal{K}^t}, z, S \big)$ \Comment{Scaling}
\State
\State {{\textsc{LocalDP}}$( \Delta_k, z, S)$}
\State \hspace{0.5cm}$ \sigma = z \cdot S $
\State \hspace{0.5cm}\textbf{return} $ \mathcal{N}(0, \sigma^2 \mathbf I) + \textsc{Clip}(\Delta_k)$ \Comment{Equations \ref{eqn_clipping}}
\State
\State \colorbox{pink}{{{\textsc{SecureSumDP}}$(\{ \Delta_k \}_{k \in \mathcal{K}},z,S)$}}
\State \hspace{0.5cm}$ \sigma = z \cdot S $
\State \hspace{0.5cm}\textbf{return} $ \mathcal{N}(0, \sigma^2 \mathbf I) + \sum_{k \in \mathcal{K}} \textsc{Clip}(\Delta_k)$ \Comment{Equations \ref{eqn_clipping}}
\State
\State {\textsc{Train}}$(t, C, W, z, S)$
\State \hspace{0.5cm}\textbf{parameters: } No. of epochs: $E$, batch size: $\beta$, lr: $\eta$
\State \hspace{0.5cm}\textbf{if} $t \leq T^{in}$: $i = 0$ \textbf{else}: $i = F_k(W)$ \Comment{Equations \ref{eqn_empirical_loss}}
\State \hspace{0.5cm}$W^{+}_{i} \leftarrow W_{i}$
\State \hspace{0.5cm}\textbf{for} each local epoch $e = 1, 2, \ldots, E$ \textbf{do}
\State \hspace{1.0cm}$\mathcal{B} \leftarrow$ (split local data into batches of size $\beta$)
\State \hspace{1.0cm}\textbf{for} batch $b \in \mathcal{B}$ \textbf{do}
\State \hspace{1.5cm}$W^{+}_{i} \leftarrow W^{+}_{i} - \eta \triangledown F_k( W^{+}_{i} )$
\State \hspace{0.5cm}\textbf{if} $t \leq T^{in}$
\State \hspace{1.0cm}\textbf{return} $\textsc{LocalDP} \big( \textsc{FILTER} \big( W^{+}_{i} - W_{i} \big), z, S \big)$
\State \hspace{0.5cm}\textbf{else}
\State \hspace{1.0cm}\textbf{return} $W^{+} - W$
\end{algorithmic}
\label{alg_pina}
\end{algorithm}
\vspace*{-10pt}
\end{figure}

In this work, we use pre-trained transformer-based models, well-suited for cross-device FL due to their ability to leverage large-scale server data for pre-training and efficient client-side fine-tuning, enabling robust performance on heterogeneous data~\cite{fl_from_pretrained, fl_large_model, dp-dylora, vit_fl}.

\subsection{Cluster model initialization}
\label{sec:initialization}

For a number of initial FL rounds $T^{in}$, clients $\mathcal{K}^t \subseteq \mathcal{K}$ sampled at round $t \leq T^{in}$ apply LoRA with $r=1$ to the same pre-trained weights as in Equation~\ref{eqn_lora}, freeze all weights except for the value projection matrix of the last attention layer denoted by $W_v^l$, and train on local data. This leads to the number of trainable parameters for the initialization stage being equal to the hidden size $h$. Each sampled client then applies LDP to the two largest positive/negative values from the model updates both in terms of absolute value, with all the other values converted to zeros. When applying LDP, we set the clipping threshold $S^{in}$ for stage 1 by:
\begin{equation}
S^{in} = \sqrt{n \left( \sqrt{\frac{(S)^2}{h}} \right)^2} = \sqrt{\frac{n}{h}} S,
\label{eqn_init_clip_norm}
\end{equation}
where $n$ is the number of non-zero values and $S$ is the clipping threshold for stage 2.
These vectors with four non-zero values are then shared directly with the server, which runs a clustering algorithm (e.g. $k$-means) to obtain initial cluster models to be trained in the second stage.

Here, we minimize the number of non-zero values transmitted to the server since the server has to have access to individual model updates to perform clustering. LDP is therefore required to privatized the shared updates, which will be too noisy to give meaningful clustering results with a large number of non-zero values being shared. However, updates to four parameters provide limited information. It is therefore also necessary to only train a small number of parameters for this step to maximize the ratio of the number of updates received by the server to the total number of trainable parameters.

\begin{table*}[!t]
  \caption{Results and comparisons on rotated CIFAR-10, rotated FMNIST and FEMNIST. We boldface the best accuracy.}
  \label{tab:main}
  \centering
  \setlength{\tabcolsep}{5pt}
  \begin{tabular}{@{}lccccccccc@{}}
    \toprule
                & \multicolumn{3}{c}{Rotated CIFAR-10} & \multicolumn{3}{c}{Rotated FMNIST} & \multicolumn{3}{c}{FEMNIST} \\ \cmidrule(lr){2-4} \cmidrule(lr){5-7} \cmidrule(lr){8-10}
    Method      & $\epsilon=\infty$    & $\epsilon=2$     & $\epsilon=8$     & $\epsilon=\infty$    & $\epsilon=2$     & $\epsilon=8$   & $\epsilon=\infty$    & $\epsilon=2$     & $\epsilon=8$ \\
    \midrule
    FedAvg                & 92.1$\pm$0.10     & 90.2$\pm$0.28     & 90.5$\pm$0.15     & 87.8$\pm$0.11     & 86.0$\pm$0.05     & 86.2$\pm$0.05     & 81.1$\pm$0.20     & 79.2$\pm$0.20     & 79.4$\pm$0.11 \\
    FedProx ($\mu=0$)    & 92.3$\pm$0.28     & 90.4$\pm$0.20     & 90.6$\pm$0.20     & 87.8$\pm$0.10     & 86.1$\pm$0.10     & 86.2$\pm$0.10     & 80.8$\pm$0.11     & 79.2$\pm$0.05     & 79.3$\pm$0.11 \\
    FedProx ($\mu=1$)    & 92.2$\pm$0.26     & 90.3$\pm$0.20     & 90.6$\pm$0.10     & 87.9$\pm$0.10     & 86.0$\pm$0.20     & 86.2$\pm$0.05     & 81.0$\pm$0.05     & 79.4$\pm$0.10     & 79.5$\pm$0.20 \\
    FedNova              & 90.3$\pm$0.10     & 89.7$\pm$0.41     & 89.9$\pm$0.23     & 88.8$\pm$0.45     & 86.9$\pm$1.21     & 87.0$\pm$0.97     & 80.4$\pm$0.37     & 79.5$\pm$0.15     & 80.0$\pm$0.60 \\
    IFCA                    & \textbf{94.3$\pm$0.20}     & 34.4$\pm$2.26     & 73.8$\pm$2.54     & 89.4$\pm$0.46     & 21.4$\pm$1.15     & 64.3$\pm$3.20     & 81.7$\pm$1.19     & 4.6$\pm$0.25     & 5.5$\pm$0.37 \\
    \midrule
    PINA (Ours)         & 93.8$\pm$0.30     & \textbf{92.6$\pm$0.25}     & \textbf{92.9$\pm$0.15}     & \textbf{89.9$\pm$0.10}     & \textbf{88.9$\pm$0.28}     & \textbf{89.0$\pm$0.23}     & \textbf{83.7$\pm$0.10}     & \textbf{82.3$\pm$0.30}     & \textbf{82.5$\pm$0.11} \\
    \bottomrule
    \end{tabular}
\end{table*}

\vspace{-1pt}
\subsection{Cluster model training}
\vspace{-1pt}

\begin{figure}[!t]
\centering
\includegraphics[width=1.0\linewidth]{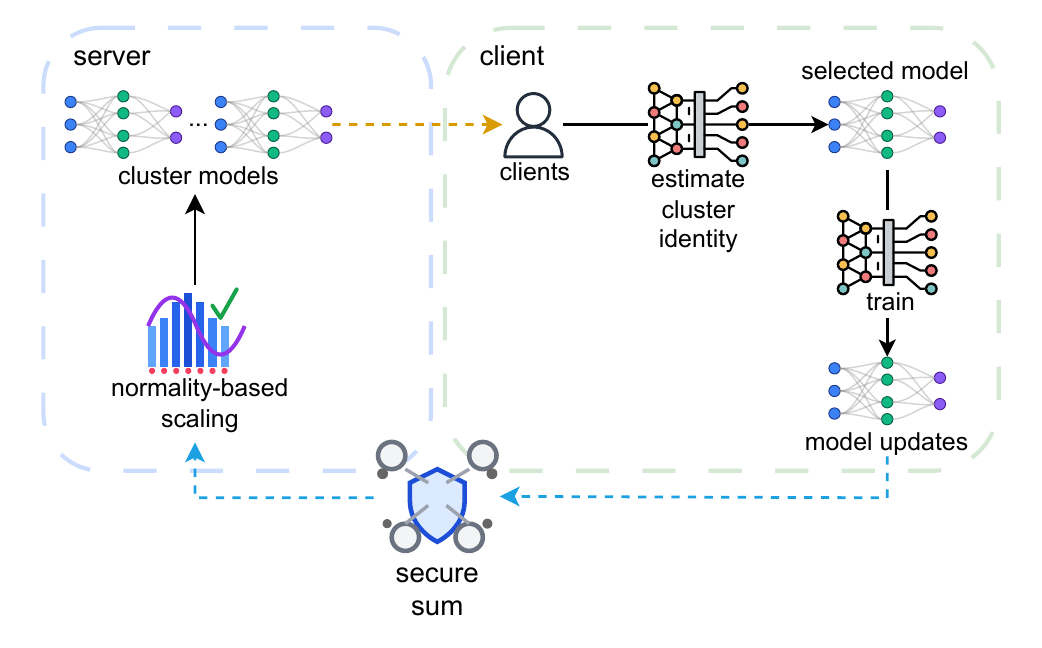}
\vspace*{-25pt}
\caption{Stage 2 of \textit{PINA} with normality-based update scaling.}
\label{fig_update_all}
\vspace*{-7pt}
\end{figure}

After initializing global cluster models, server sends the latest cluster models to each sampled client at the start of each communication round for training. As in \cite{ifca}, clients perform cluster identification based on training loss and train the selected cluster model on local data.
Let $\mathcal{Z} = \{z_1, \ldots, z_n\}$ be the samples held by client $k \in \mathcal{K}$.
We define the empirical loss $F$ associated with client $k$ as follows:
\begin{equation}
F_k(W) = \frac{1}{|\mathcal{Z}|} \sum_{z \in \mathcal{Z}} f(W;z),
\label{eqn_empirical_loss}
\end{equation}
where $f(W;z)$ is the loss function associated with sample $z$.

Clients then clip model updates to a predefined threshold $S$ by:
\begin{equation}
\textsc{Clip}(\Delta_k) = \Delta_k \cdot \min\big(1, \frac{S}{\lVert \Delta_k \rVert_2} \big),
\label{eqn_clipping}
\end{equation}
and share the clipped updates with the server. The updates are then aggregated and noised before being applied to the global cluster models. The aggregation and noise addition can be implemented via a secure sum protocol such as SecAgg \cite{secagg}, where each client adds noise to their clipped updates known as distributed DP (DDP) \cite{secure_sum_2}, ensuring the aggregated updates have the correct noise \cite{secure_sum}. The server receives only the noisy, aggregated updates, with no access to individual client contributions, providing privacy guarantees even against an untrusted server. As secure aggregation is orthogonal to our core contribution, we abstract it (highlighted in pink in Algorithm~\ref{alg_pina}) as a black-box layer.

After receiving the aggregated noisy updates through secure sum, for a number of communication rounds $T^{no}$, the server normalizes the magnitude of each aggregate $\Delta_i$ to the one with the smallest $\ell_2$ norm by:
\begin{equation}
\Delta_i = \Delta_i \cdot \frac{\lVert \Delta_{\hat{i}} \rVert_2}{\lVert \Delta_i \rVert_2},
\label{eqn_update_normalization}
\end{equation}
where $i \in [\![1,C]\!]$ and $\hat{i} = \text{argmin}_{j \in [\![1,C]\!]} \big( \lVert \Delta_j \rVert_2 \big)$ with $[\![1,C]\!] = \{1,\ldots,C\}$. This is done to stabilize early training of cluster models.

After round $T^{in}+T^{no}$, for the remainder of training, we alternatively scale the aggregated model updates to each global cluster model based on its normality, estimated via the Shapiro-Wilk test statistic~\cite{shapiro} as follows:
\begin{equation}
\Delta_i = \begin{cases}
\Delta_i \cdot \frac{w_i}{\sum_{j \in [\![1,C]\!]} w_j}, & \text{if } w_i < 0.99
\\
\Delta_i \cdot 0, & \text{otherwise}
\end{cases},
\label{eqn_update_scaling_shapiro}
\end{equation}
where $w_i$ denotes the Shapiro-Wilk test statistic for the $i$th cluster's aggregated updates. Since each client shares updates to all cluster models, including the ones that are not selected and trained, clusters chosen by only a few clients experience significantly reduced update magnitudes after aggregation, leading to slow convergence. To address this, we propose this novel scaling mechanism, which restores the intended $\ell_2$ norm of the updates to global cluster models. This improves both the robustness and fairness of the proposed framework. To avoid applying amplified noise to a global cluster model with zero contributing clients, we zero out the update whenever the test statistic reaches the threshold of $0.99$.

\section{Experiments}
\label{sec:experiments}

\textbf{Experimental settings:} We use privacy budget of $\epsilon \in \{2,8\}$ which are commonly used in existing works \cite{dp-fedprox, dp-scaffold} and $\delta=\frac{1}{|\mathcal{K}|^{1.1}}$ \cite{dp-fedavg}.
Following \cite{dp-fedavg, flair, dp-dylora}, we simulate a cohort size of 10k with a smaller cohort size to achieve a more realistic signal-to-noise ratio which represents industry scale more closely.
We use rotated CIFAR-10 ($C = 2$), rotated FMNIST ($C = 4$) and FEMNIST ($C = 2$) for our experiments. Following \cite{ifca}, we generate the first two by applying the same rotation (0, 180 degrees for CIFAR-10 and 0, 90, 180, 270 degrees for FMNIST) to all images of a client. We set the total number of clients to 5,000 for CIFAR-10/FMNIST and 2,840 for FEMNIST.
We use standard data augmentation (e.g. random cropping and horizontal flipping) to increase the size of the training set by 5x for CIFAR-10 and FMNIST. For all experiments, clients train for $E=10$ local epochs with a batch size of $\beta=50$ and learning rate of $\eta=0.01$.

We compare PINA with state-of-the-art (SOTA) methods including FedAvg \cite{fedavg}, FedProx \cite{fedprox}, FedNova \cite{fednova} and IFCA \cite{ifca}.
For IFCA, we set the clipping threshold to the same value as our method and apply LDP for $\epsilon \in \{2, 8\}$.
It was shown in \cite{ifca} that IFCA could be combined with personalized FL to improve performance even further. In this paper, we do not consider personalized FL.
We exclude SCAFFOLD \cite{scaffold} from our experiments since it is designed for the cross-silo setting \cite{scaffold_cross_silo} and we only focus on the more challenging cross-device setting in this work. Client dropouts are not considered in our experiments.
We use a 22M-parameter ViT-Small model pre-trained on ImageNet-21k and publicly available via Hugging Face. We assume all clients are provided with this pre-trained model at the start of training, consistent with standard FL practices.

\textbf{Comparing with existing DP-FL methods:} Table~\ref{tab:main} shows results for different methods with non-private FL ($\epsilon=\infty$) and DP-FL ($\epsilon \in \{2, 8\}$). The results demonstrate that PINA consistently outperforms SOTA methods across all three datasets. In particular, for $\epsilon = 2$, PINA achieves up to $2.9\%$, $2.9\%$ and $3.1\%$ improvements on rotated CIFAR-10, rotated FMNIST and FEMNIST, respectively. For $\epsilon = 8$, the improvements reach up to $3.0\%$, $2.8\%$ and $3.2\%$ on the same datasets. Notably, the improvements are most pronounced on the naturally non-IID dataset FEMNIST, highlighting the ability of our proposed method to handle real-world data heterogeneity. Overall, PINA shows an average accuracy improvement of $2.9\%$ over SOTA methods for $\epsilon \in \{2, 8\}$ on non-IID data, validating its effectiveness in learning on heterogeneous user data while preserving user privacy.

\begin{figure}[!h]
\vspace*{-6pt}
    \centering
    \subfloat[{\scriptsize Clustering accuracy}]{%
        \includegraphics[width=0.5\linewidth]{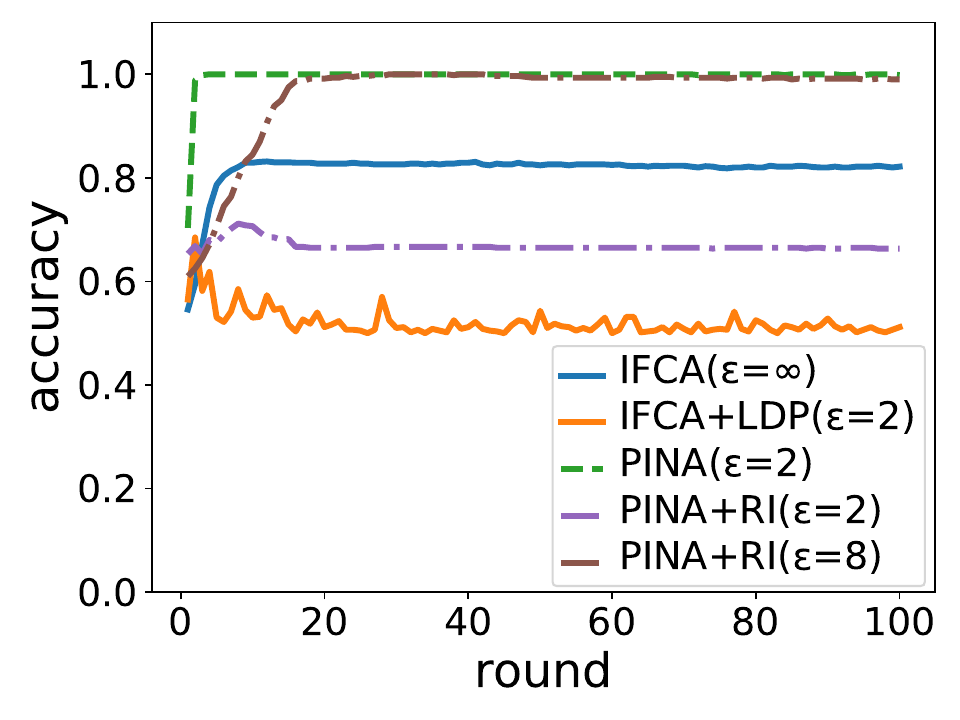}
        \label{fig_clustering_accuracy}
    }
    \subfloat[{\scriptsize Classification accuracy}]{%
        \includegraphics[width=0.5\linewidth]{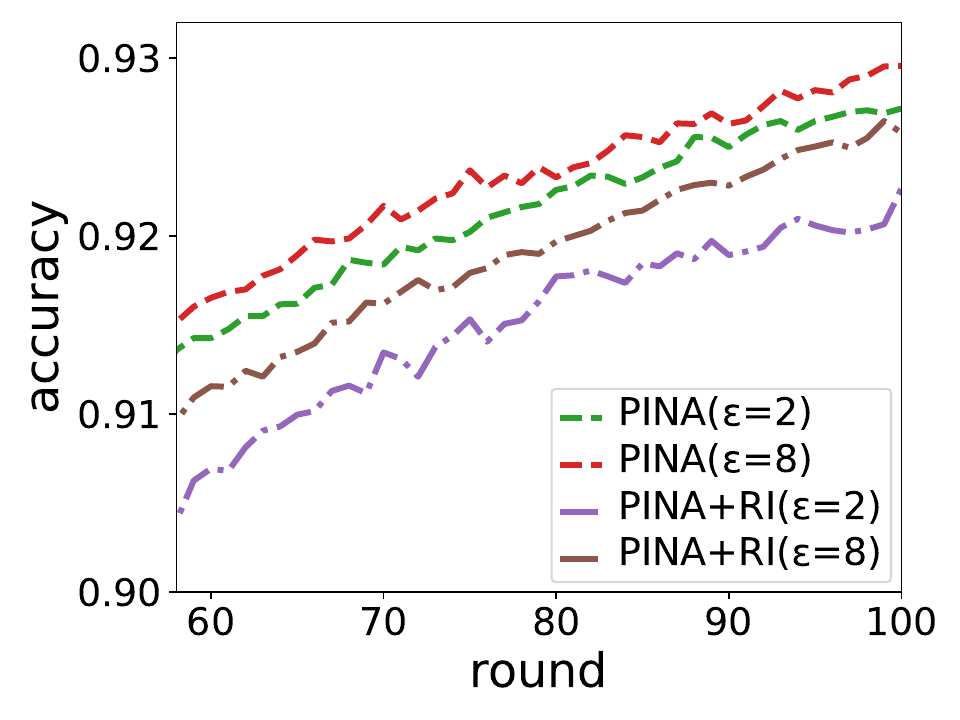}
        \label{fig_classification_accuracy}
    }
    \caption{Comparison of clustering and classification accuracy on CIFAR-10. PINA+RI denotes our proposed method with random initialization. Results are averaged over three runs.}
    \label{fig_clustering_classification_accuracy}
\vspace*{2pt}
\end{figure}

\textbf{Ablation:} Fig.~\ref{fig_clustering_accuracy} illustrates the clustering accuracy of IFCA and our proposed method under varying levels of added noise. As evident from the results, our method achieves optimal clustering with a stringent privacy budget of $\epsilon = 2$. In contrast, incorporating LDP into IFCA throughout the entire training process has a detrimental effect on the clustering structure. Notably, our method attains optimal clustering even with random initialization when the privacy budget is relaxed to $\epsilon = 8$, demonstrating its robustness. Furthermore, as shown in Fig~\ref{fig_classification_accuracy}, our novel initialization mechanism enables the algorithm to converge to the optimal clustering structure significantly faster than random initialization, resulting in accelerated convergence and improved classification accuracy.

\section{Conclusion}
\label{sec:conclusion}

In this work, we propose PINA, a privacy-preserving clustered FL framework that effectively mitigates data heterogeneity in DP-FL. By combining privatized client sketches for robust initialization and a normality-driven aggregation mechanism that accounts for imbalanced contributions, PINA achieves superior performance on non-IID data without requiring server-side privileged data or random restarts. Extensive experiments show that PINA consistently outperforms SOTA DP-FL methods under non-IID settings, achieving an average improvement of $2.9\%$ in test accuracy.


\vfill\pagebreak

\bibliographystyle{IEEEbib}
\bibliography{refs}

@inproceedings{lora,
    title={Lo{RA}: Low-Rank Adaptation of Large Language Models},
    author={Hu, Edward J and Shen, Yelong and Wallis, Phillip and Allen-Zhu, Zeyuan and Li, Yuanzhi and Wang, Shean and Wang, Lu and others},
    booktitle={ICLR},
    year={2022}
}

@inproceedings{fedavg,
  title={Communication-efficient learning of deep networks from decentralized data},
  author={McMahan, Brendan and Moore, Eider and Ramage, Daniel and Hampson, Seth and others},
  booktitle={AISTATS},
  year={2017}
}

@article{fl_heterogeneity,
  title={Federated learning: Challenges, methods, and future directions},
  author={Li, Tian and Sahu, Anit Kumar and Talwalkar, Ameet and Smith, Virginia},
  journal={IEEE signal processing magazine},
  year={2020},
}

@inproceedings{fl_large_model,
  title={Scaling language model size in cross-device federated learning},
  author={Ro, Jae and Breiner, Theresa and McConnaughey, Lara and Chen, Mingqing and Suresh, Ananda and others},
  booktitle={FL4NLP},
  year={2022}
}

@article{fl_from_pretrained,
  title={Federated learning from pre-trained models: A contrastive learning approach},
  author={Tan, Yue and Long, Guodong and Ma, Jie and Liu, Lu and Zhou, Tianyi and Jiang, Jing},
  journal={NeurIPS},
  year={2022}
}

@inproceedings{vit_fl,
  title={Rethinking architecture design for tackling data heterogeneity in federated learning},
  author={Qu, Liangqiong and Zhou, Yuyin and Liang, Paul Pu and Xia, Yingda and Wang, Feifei and others},
  booktitle={CVPR},
  year={2022}
}

@article{fedprox,
  title={Federated optimization in heterogeneous networks},
  author={Li, Tian and Sahu, Anit Kumar and Zaheer, Manzil and Sanjabi, Maziar and Talwalkar, Ameet and Smith, Virginia},
  journal={MLSys},
  year={2020}
}

@inproceedings{scaffold,
  title={Scaffold: Stochastic controlled averaging for federated learning},
  author={Karimireddy, Sai Praneeth and Kale, Satyen and Mohri, Mehryar and Reddi, Sashank and others},
  booktitle={ICML},
  year={2020}
}

@article{fednova,
  title={Tackling the objective inconsistency problem in heterogeneous federated optimization},
  author={Wang, Jianyu and Liu, Qinghua and Liang, Hao and Joshi, Gauri and Poor, H Vincent},
  journal={NeurIPS},
  year={2020}
}

@article{scaffold_cross_silo,
  title={Breaking the centralized barrier for cross-device federated learning},
  author={Karimireddy, Sai Praneeth and Jaggi, Martin and Kale, Satyen and Mohri, Mehryar and Reddi, Sashank and others},
  journal={NeurIPS},
  year={2021}
}

@article{ifca,
  title={An efficient framework for clustered federated learning},
  author={Ghosh, Avishek and Chung, Jichan and Yin, Dong and Ramchandran, Kannan},
  journal={NeurIPS},
  year={2020}
}

@article{cfl,
  title={Clustered federated learning: Model-agnostic distributed multitask optimization under privacy constraints},
  author={Sattler, Felix and M{\"u}ller, Klaus-Robert and Samek, Wojciech},
  journal={IEEE TNNLS},
  year={2020}
}

@inproceedings{pacfl,
  title={Efficient distribution similarity identification in clustered federated learning via principal angles between client data subspaces},
  author={Vahidian, Saeed and Morafah, Mahdi and Wang, Weijia and Kungurtsev, Vyacheslav and Chen, Chen and others},
  booktitle={AAAI},
  year={2023}
}

@InProceedings{dp,
    author="Dwork, Cynthia
    and McSherry, Frank
    and Nissim, Kobbi
    and Smith, Adam",
    title="Calibrating Noise to Sensitivity in Private Data Analysis",
    booktitle="Theory of Cryptography",
    year="2006"
}

@article{ldp,
  title={What can we learn privately?},
  author={Kasiviswanathan, Shiva Prasad and Lee, Homin K and Nissim, Kobbi and Raskhodnikova, Sofya and Smith, Adam},
  journal={SIAM Journal on Computing},
  year={2011},
}

@inproceedings{ddp_secure_sum,
  title={Benchmarking Secure Sampling Protocols for Differential Privacy},
  author={Fu, Yucheng and Wang, Tianhao},
  booktitle={CCS},
  year={2024}
}

@article{gaussian_mechanism,
    title={The algorithmic foundations of differential privacy},
    author={Dwork, Cynthia and Roth, Aaron and others},
    journal={Foundations and Trends{\textregistered} in Theoretical Computer Science},
    year={2014}
}

@inproceedings{moments_accountant,
    author = {Abadi, Martin and Chu, Andy and Goodfellow, Ian and McMahan, H Brendan and Mironov, Ilya and Talwar, Kunal and Zhang, Li},
    title = {Deep Learning with Differential Privacy},
    year = {2016},
    booktitle = {CCS}
}

@inproceedings{rdp,
  title={R{\'e}nyi differential privacy},
  author={Mironov, Ilya},
  booktitle={2017 IEEE 30th computer security foundations symposium (CSF)},
  year={2017}
}

@inproceedings{rdp_accountant,
  title={Hypothesis testing interpretations and {R}enyi differential privacy},
  author={Balle, Borja and Barthe, Gilles and Gaboardi, Marco and Hsu, Justin and Sato, Tetsuya},
  booktitle={International Conference on Artificial Intelligence and Statistics},
  year={2020},
  organization={PMLR}
}

@inproceedings{privacy_amplification_via_subsampling_1,
    author = {Balle, Borja and Barthe, Gilles and Gaboardi, Marco},
    booktitle = {NeurIPS},
    title = {Privacy Amplification by Subsampling: Tight Analyses via Couplings and Divergences},
    year = {2018}
}

@article{dp-fedavg,
  title={Learning differentially private recurrent language models},
  author={McMahan, H Brendan and Ramage, Daniel and Talwar, Kunal and Zhang, Li},
  journal={ICLR},
  year={2018}
}

@article{dp-fedprox,
  title={Federated learning with differential privacy: Algorithms and performance analysis},
  author={Wei, Kang and Li, Jun and Ding, Ming and Ma, Chuan and Yang, Howard H and others},
  journal={IEEE TIFS},
  year={2020}
}

@inproceedings{dp-scaffold,
  title={Differentially private federated learning on heterogeneous data},
  author={Noble, Maxence and Bellet, Aur{\'e}lien and Dieuleveut, Aymeric},
  booktitle={AISTATS},
  year={2022}
}

@article{flair,
  title={{FLAIR}: Federated learning annotated image repository},
  author={Song, Congzheng and Granqvist, Filip and Talwar, Kunal},
  journal={NeurIPS},
  year={2022}
}

@ARTICLE{ifca_ldp,
  author={He, Zaobo and Wang, Lintao and Cai, Zhipeng},
  journal={IEEE IoT}, 
  title={Clustered Federated Learning With Adaptive Local Differential Privacy on Heterogeneous {IoT} Data}, 
  year={2024}
}

@article{r-dpcfl,
  title={Mitigating disparate impact of differential privacy in federated learning through robust clustering},
  author={Malekmohammadi, Saber and Taik, Afaf and Farnadi, Golnoosh},
  journal={arXiv},
  year={2024}
}

@inproceedings{sacfl,
  title={Secure aggregation for clustered federated learning},
  author={Sami, Hasin Us and G{\"u}ler, Ba{\c{s}}ak},
  booktitle={ISIT},
  year={2023}
}

@article{clusterguard,
  title={ClusterGuard: Secure Clustered Aggregation for Federated Learning with Robustness},
  author={Zhao, Yulin and Wan, Zhiguo and Guan, Zhangshuang and others},
  journal={Cryptology ePrint Archive},
  year={2024}
}

@inproceedings{dpfl_heterogeneity,
  title={$\{$PrivateFL$\}$: Accurate, differentially private federated learning via personalized data transformation},
  author={Yang, Yuchen and Hui, Bo and Yuan, Haolin and Gong, Neil and Cao, Yinzhi},
  booktitle={USENIX Security},
  year={2023}
}

@article{dp-dylora,
  title={{DP}-{D}y{L}o{RA}: Fine-Tuning Transformer-Based Models On-Device under Differentially Private Federated Learning using Dynamic Low-Rank Adaptation},
  author={Xu, Jie and Saravanan, Karthikeyan and van Dalen, Rogier and Mehmood, Haaris and Tuckey, David and Ozay, Mete},
  journal={arXiv preprint arXiv:2405.06368},
  year={2024}
}

@inproceedings{secure_sum_2,
  title={A hybrid approach to privacy-preserving federated learning},
  author={Truex, Stacey and Baracaldo, Nathalie and Anwar, Ali and Steinke, Thomas and Ludwig, Heiko and Zhang, Rui and Zhou, Yi},
  booktitle={AISec},
  year={2019}
}

@inproceedings{secagg,
    author    = {Bonawitz, Keith and Ivanov, Vladimir and Kreuter, Ben and Marcedone, Antonio and others},
    title     = {Practical Secure Aggregation for Privacy-Preserving Machine Learning},
    year      = {2017},
    booktitle = {CCS}
}

@article{secure_sum,
    author  = {Goryczka, Slawomir and Xiong, Li},
    journal = {IEEE Transactions on Dependable and Secure Computing},
    title   = {A Comprehensive Comparison of Multiparty Secure Additions with Differential Privacy},
    year    = {2017}
}

@article{shapiro,
  title={An analysis of variance test for normality (complete samples)},
  author={Shapiro, Samuel Sanford and Wilk, Martin B},
  journal={Biometrika},
  year={1965},
}

\end{document}